\documentclass[letterpaper, 10 pt, conference]{ieeeconf}  % Comment this line out
\IEEEoverridecommandlockouts                              % This command is only
\usepackage[utf8]{inputenc}
\usepackage[T1]{fontenc}
\usepackage{graphicx}
\usepackage{subcaption}
\usepackage{xcolor}
\usepackage{cite}

\title{\LARGE \bf
Classification of Safety Driver Attention During Autonomous Vehicle Operation
}

\author{Santiago Gerling Konrad, Julie Stephany Berrio, Mao Shan, Favio Masson, Eduardo Nebot and Stewart Worrall%
\thanks{This work has been supported by the Australian Centre for Robotics (ACFR). The authors are with the ACFR at the University of Sydney (NSW, Australia). E-mails: {\{j.berrio, m.shan, e.nebot, s.worrall}\}@acfr.usyd.edu.au; and IIIE/DIEC Universidad Nacional del Sur-CONICET (Bahía Blanca, Argentina), E-Mails: {\{santiago.konrad, fmasson}\}@uns.edu.ar.}
}

\begin{document}

\maketitle
\thispagestyle{empty}
\pagestyle{empty}

%%%%%%%%%%%%%%%%%%%%%%%%%%%%%%%%%%%%%%%%%%%%%%%%%%%%%%%%%%%%%%%%%%%%%%%%%%%%%%%%
\begin{abstract}
Despite the continual advances in Advanced Driver Assistance Systems (ADAS) and the development of high-level autonomous vehicles (AV), there is a general consensus that for the short to medium term, there is a requirement for a human supervisor to handle the edge cases that inevitably arise.
Given this requirement, it is essential that the state of the vehicle operator is monitored to ensure they are contributing to the safe operation of the vehicle.
This paper introduces a dual-source approach integrating data from an infrared camera facing the vehicle operator and vehicle perception systems to produce a metric for driver alertness in order to promote and ensure safe operator behavior. The infrared camera detects the driver's head, enabling the calculation of head orientation which is relevant as the head typically moves according to the individual's focus of attention. By incorporating environmental data from the perception system, it becomes possible to determine whether the vehicle operator is observing objects in the surroundings.
Experiments were conducted using data collected in Sydney, Australia, simulating AV operations in an urban environment. Our results demonstrate that the proposed system effectively determines a metric for the attention levels of the vehicle operator, enabling interventions such as warnings or reducing autonomous functionality as appropriate. This comprehensive solution shows promise in contributing to ADAS and AVs' overall safety and efficiency in a real-world setting.
\end{abstract}

\section{Introduction}

As the prevalence of Advanced Driver Assistance Systems (ADAS) and Autonomous Vehicles (AVs) increases, monitoring the vehicle operator's state becomes essential to ensure safety. AVs of level 3 or below require human intervention when necessary, and the safety driver must remain alert to monitor the vehicle's surroundings. In a driving environment, the surrounding elements, including pedestrians, vehicles, and infrastructure, are perceived by the vehicle sensors, and play a crucial role in driving planning and navigation. The safety driver monitoring the driving process must be trained to pay attention to these agents to ensure overall safety. 

The attention level is a function of the topology of the road, the type and number of traffic participants, and the dynamic and unpredictable nature of the environment, which makes each situation unique. Maintaining constant attention to the environment is a great challenge for the safety driver, and sometimes may not even be possible due to the number of simultaneous events that need to be monitored. Therefore, in similar traffic situations, driver attention levels may vary. 

Tracking the driver state can help ensure that the driver is aware of their environment and capable of taking control of the vehicle when required.
Driver alertness is critical in maintaining the safety and efficiency of AVs. While traditional Driver Monitoring Systems (DMS) focus primarily on driver information, an advanced system incorporating vehicle and surrounding data can provide a more comprehensive understanding of driver behaviour.

This paper addresses the challenge of identifying inattentive human drivers in the context of ADAS and AVs by introducing a dual-source approach that combines both driver and vehicle data. We consider the objects detected by the vehicle's perception system and the observations made by the driver.
This approach provides a metric to estimate driver alertness, enabling future warnings and interventions that would promote safe driving behaviours, contributing to these emerging technologies' overall safety and efficiency in a single, comprehensive solution.

To validate our approach, we conducted data collection during real driving scenarios. We utilized an IR camera to capture images of the safety driver, allowing for the calculation of head orientation. Additionally, we logged information about the surrounding environment captured by the vehicle sensors, including objects such as pedestrians and vehicles. The fusion of data from both sources enabled the classification of driver attention into two levels: regular and low. Based on our experiments, we have observed that the driver's attention gradually decreases over time as they become more accustomed and confident with the technology. This phenomenon is often referred to as ``automation complacency'' or ``attentional disengagement''. Research in the field has shown that as individuals gain trust in automated systems, they may become less vigilant and attentive to their surroundings. These findings indicate that our approach is capable of measuring the driver's level of attention towards their surrounding environment.

\section{Related Work}

A comprehensive safety case for ADAS and AVs should incorporate a trained human supervisor capable of responding promptly to any autonomy failure \cite{Eriksson2017-1, Eriksson2017-2, FlynnEvans2021, Pipkorn2022}. Additionally, the safety case must include an autonomy failure profile compatible with adequate human supervision \cite{Koopman2019}, and the human supervisor must be able to manage any autonomy failures.

Drivers are typically unable to predict when they may be required to assume control of an AV. A lack of alertness could impede their ability to respond effectively in emergencies. To enhance safety in AVs, several studies have been conducted to examine the underlying factors that contribute to poor driver alertness. In \cite{FlynnEvans2021, Tran2021, Cunningham2018}, the authors investigate some primary factors contributing to driver negligence. The main factors are passive fatigue, distraction, over-reliance on automation, and prolonged driving time. A Human Supervisor Monitoring System (HSMS) can improve road safety by constantly monitoring the presence and state of the human supervisor, providing real-time alerts to intervene if necessary, and ensuring that supervisors are always prepared to take control of the vehicle when needed.

In recent studies \cite{Amina2020, Kumar2018}, three main approaches to detecting driver's state have been identified and categorized as physiological, behavioural, and vehicle-based. The physiological approach involves using sensors attached to the human body to collect signals such as ECG, PPG, EEG, EOG, skin temperature, GSR, and EMG. While effective, this method can be expensive and intrusive, and signals can vary between individuals. The behavioural approach involves monitoring a driver's behaviour for signs of fatigue or distraction, such as head pose, blink frequency, PERCLOS, and gaze region. While non-intrusive, this approach faces challenges such as facial occlusion and fast head movements. The vehicle-based approach is a reliable and non-intrusive method that utilizes sensors in various vehicle components to collect data on metrics such as steering wheel movement, acceleration, braking, geo-position, and object detection. This approach can be limited in detecting certain types of driver states or behaviours, such as fatigue or distraction that may not manifest as changes in vehicle movement.

In \cite{Friedrichs2020}, a combination of the behaviour and vehicle-based approaches was utilized. However, determining the overall rate of observed failures involves the product of the autonomy failure rate and the rate of unsuccessful failure mitigation by the supervisor, as noted in \cite{Koopman2019}. One of the challenges is that human ability varies in a non-linear manner with autonomy failure rates, making it more challenging for a supervisor to ensure safety as autonomy maturity improves. Therefore, safety cases for road testing must consider both the anticipated failures during testing and the practical efficacy of human supervisors, given the failure profile.

Moreover, research studies have shown that as drivers gain trust and confidence in the capabilities of autonomous vehicles, their attention to the driving task may diminish over time \cite{JAMSON2013116}. Factors contributing to automation complacency include the perception of increased system reliability, a sense of reduced workload, and a perception that the system can handle challenging driving situations effectively \cite{mental}.

This paper introduces a non-invasive driver monitoring system that combines sensor information from the driver and vehicle to detect inattention patterns and generate a metric of driver attention that can be used for warnings and possible interventions. The system primarily focuses on objects in the driver's immediate vicinity, which are critical to safe driving. Through experiments using real-world data, we have verified that our proposed metric aligns with well-researched studies on automation complacency.
The system does not account for fatigue or drowsiness and aims to address only a lack of attention to the driving environment.

\section{Methodology}

The proposed system assesses driver alertness using a combination of driver, vehicle, and environmental information. In driving scenarios, the vehicle environment constantly changes, and the vehicle must handle complex situations such as approaching an intersection, navigating traffic lights, or interacting with pedestrians and other vehicles. The human supervisor must monitor these situations; head movement variations do not necessarily indicate inattentiveness.

\subsection{Dataset Collection}

\begin{figure}[h]
  \centering
  \includegraphics[width=0.95\columnwidth]{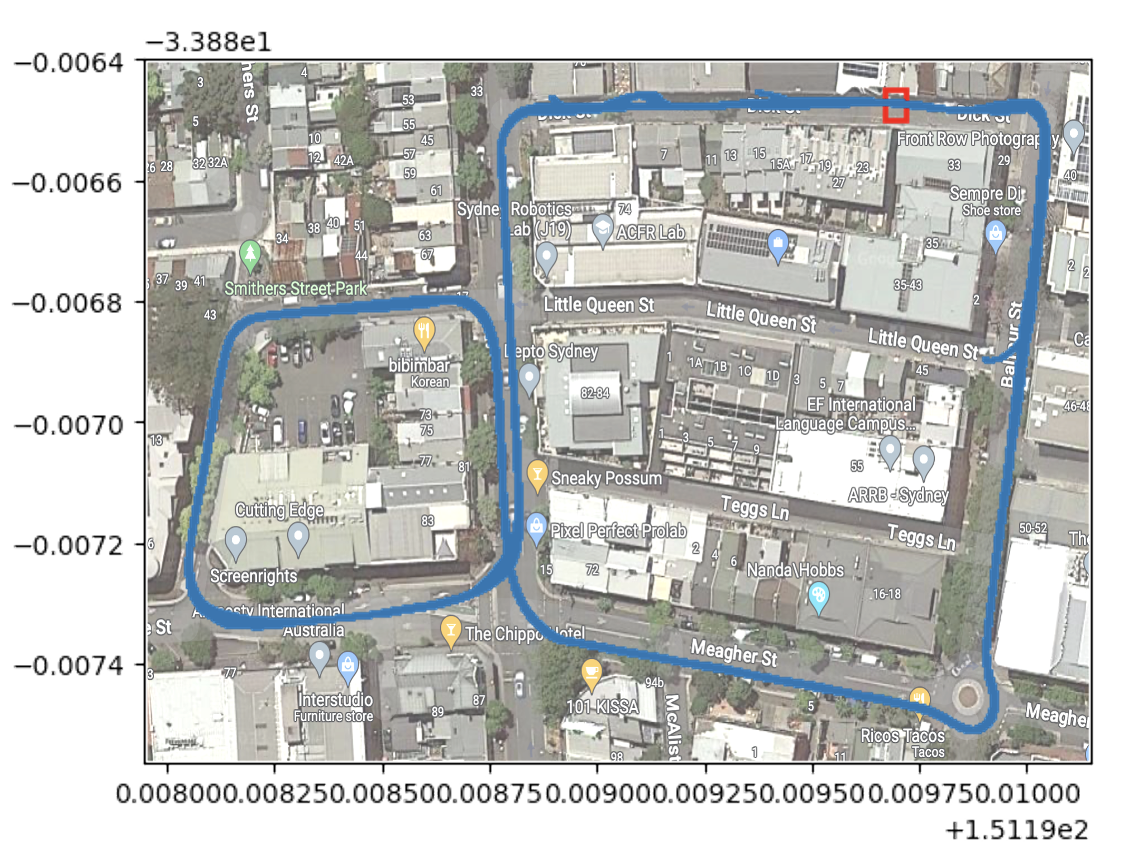}
  \caption{Path driven in UTM coordinates.}
  \label{fig:path}
\end{figure}

This study gathered data using our electric vehicle platforms, as described in \cite{usyd_dataset}. The vehicles were equipped with advanced autonomous vehicle sensors, including multiple cameras perceiving the surroundings, an interior IR camera focused on the safety driver, lidar, GPS, and odometry. These sensors were synchronized and calibrated to ensure accurate data collection. The participants in the study were simulating the role of safety drivers for an AV, providing a realistic context for data collection. The specific trajectory followed by the vehicle is illustrated in Fig \ref{fig:path}. A total of 25 laps were completed as the seven participants took turns acting as the supervisor of the vehicle along the designated path during multiple laps. The starting and endpoints of each lap are indicated by the red square depicted in Fig \ref{fig:path}.

%\subsection{Dataset Post-Processing}

%The data underwent post-processing to analyse the correlation between the detected safety driver's face direction and the outcome of the vehicle's perception system. This analysis allowed for a deeper understanding of the driver's behavior and the vehicle's surrounding environment during the drive.

\subsection{Head Pose Estimation} \label{sec:HeadOrientation}

The safety driver is monitored by an IR camera installed inside the vehicle. Using an IR camera proves advantageous in capturing clear images, particularly in scenarios where partial face illumination might hinder accurate detection. Moreover, IR cameras facilitate the identification of eye points even when individuals wear sunglasses, expanding its applicability to a broader range of cases.
This study employed a pre-trained YOLOv7-tiny model \cite{wang2022} to detect the safety driver's head using 128x128px IR images. 
Geometric models are a widely used method for image-based head pose estimation. These models utilize a static template and facial landmarks to determine the corresponding head pose through an analytical process. The main challenge of these models lies in accurately detecting the facial landmarks.

\begin{figure}[h]
  \centering
  \includegraphics[width=0.5\columnwidth]{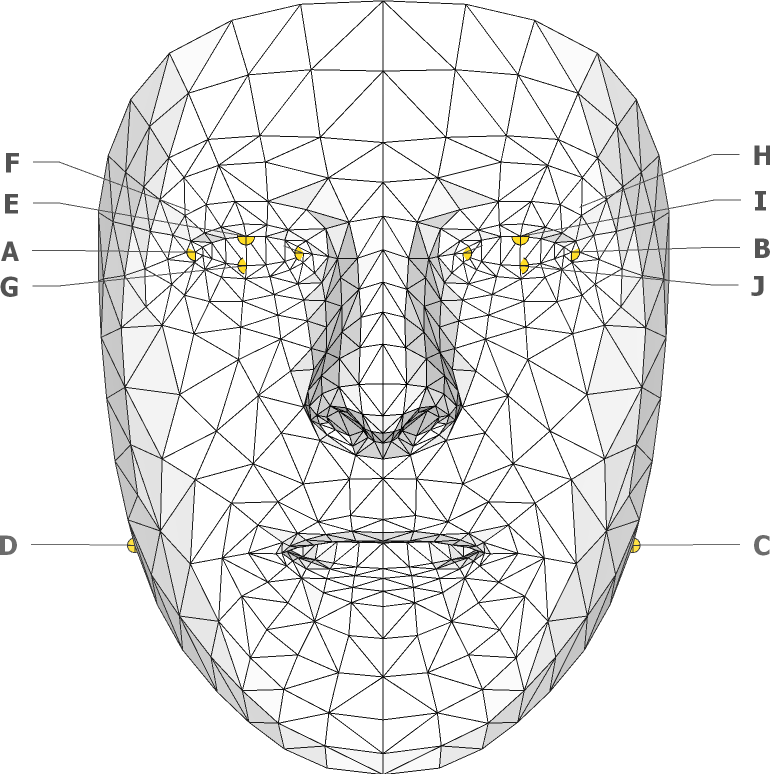}
  \caption{\small Landmarks used for head pose estimation.}
  \label{fig:face-landmarks-model}
\end{figure}

Face landmarks were estimated using MediaPipe \cite{Mediapipe2019-1, Mediapipe2019-2}, a machine learning-based method that infers the 3D facial surface from a single camera input in real-time without the need for a dedicated depth sensor. The method produces ten 3D face landmarks, which help determine the head pose.  
Fig. \ref{fig:face-landmarks-model} depicts the face landmarks/points detected.

The estimation of the head pose is achieved through the utilization of the Perspective-n-Point (PnP) algorithm that incorporates Infinitesimal Plane-based Pose Estimation (IPPE) \cite{Collins2014}. The algorithm requires that the object points be co-planar for successful implementation. For this reason, we utilized four landmarks corresponding to the outer eye's canthus and jaw angles (points A, B, C, and D) as indicated in Fig. \ref{fig:face-landmarks-model}.
The pitch, roll, and yaw angles can define the head orientation in 3D space. However, in our approach, only the yaw angles are used.

\subsection{Vehicle Perception}

The vehicle perception system is based on sensor fusion between camera and lidar information. By utilizing the extrinsic calibration between the cameras and lidar, as well as the intrinsic parameters of the cameras, the vehicle can accurately detect surrounding objects. A YOLOv4-tiny \cite{article_yolo} detector identifies pedestrians and vehicles within image frames from the three front-facing cameras. Each detection is then translated into the point cloud domain using rigid transformation matrices between sensors. Subsequently, the corresponding point cloud is clustered to extract the centroid. To track and smooth the paths of multiple objects in a 2D top-down view, a Gaussian Mixture Probability Hypothesis Density (GMPHD) filter \cite{paper:VoMa2006} is employed.

%\textcolor{red}{The acquired data lacks map information, which restricts the positioning of objects solely to their relative coordinates with respect to the vehicle. This limitation leads to a lack of comprehensive understanding of the environment, resulting in detected objects lacking contextual information. This presents challenges in identifying parked or stationary vehicles, pedestrians on sidewalks, or pedestrians crossing the street because it is not possible to differentiate the pedestrian areas or the road.
%Furthermore, the vehicle lacks information about occluded objects, such as a vehicle obstructed by another vehicle in front of the ego-vehicle or pedestrians hidden behind vehicles not within the vehicle's line of sight.}

\subsection{Information Overlay}

The data underwent post-processing to analyse the correlation between the detected safety driver's face direction and the outcome of the vehicle's perception system. This analysis allowed for a deeper understanding of the driver's behavior and the vehicle's surrounding environment during the drive.

\begin{figure}[h]
  \centering
  \includegraphics[width=0.93\columnwidth]{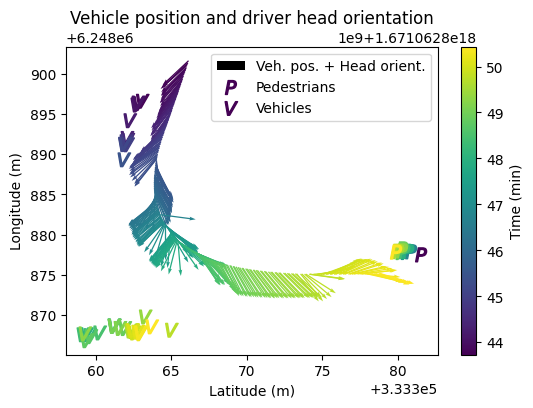}
  \caption{\small Head orientation and vehicle perception output in relation to the vehicle's position. ``V'' and ``P'' denote the positions of vehicles and pedestrians. The temporal scale is represented by a color gradient, with blue indicating the earliest data, transitioning through green and finally to yellow for the most recent data.}
  \label{fig:position}
\end{figure}

For each timestamp, we convert the head orientation (head yaw) and perception results to the local coordinate system of the ego vehicle. This transformation is illustrated in Figure \ref{fig:position}, where vehicles and pedestrians detected by the perception system are represented by the letters ``V'' and ``P'', respectively. The arrows displayed along the vehicle's pose indicate the orientation of the safety driver's head. This representation enabled clear visualization of the detected objects in relation to the ego vehicle's position and orientation over time.

\subsection{Filtering} \label{sec:filtering}

At each timestamp, the objects surrounding the vehicle were filtered to ensure that the safety driver's field of view spanned $90^{\circ}$, with $45^{\circ}$ coverage on each side. The object detection range was established at 15 m in front of the vehicle and 10 m on either side. This configuration allowed for an optimal balance between visibility and focus for the safety driver.
The head orientation estimation algorithm demonstrates limitations, such as losing head detection due to extreme head rotations or lighting conditions. It can also be affected by occlusion or partial head visibility. These factors can cause the algorithm to fail to detect the head or provide an inaccurate estimate of its orientation.

\begin{figure}[h]
\vspace{2mm}
  \centering
  \includegraphics[width=0.95\columnwidth]{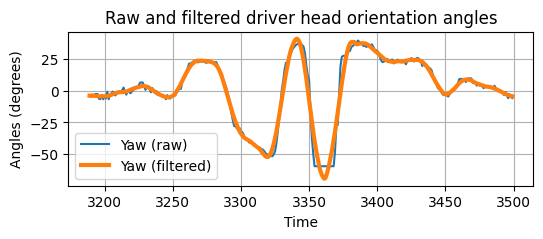}
  \caption{\small Head orientation filtering. The yaw angle of the driver's head is shown in blue, while the resulting signal of the filtering process is depicted in orange.}
  \label{fig:filtering}
\end{figure}

A filtering process is applied to the collected data to address these challenges. It identifies and eliminates noise and outliers, which are data points significantly different from most of the data. By removing these outliers, the filtered data provides a more accurate representation of the safety driver's head orientation during the simulation. Fig. \ref{fig:filtering} demonstrates the effectiveness of the filtering process by comparing the unfiltered data with the filtered data. The figure clearly shows how the filtering process improves the accuracy of the head orientation estimation by removing outliers and reducing the impact of noise in the data.

\begin{figure}[h]
  \centering
  \includegraphics[width=0.95\columnwidth]{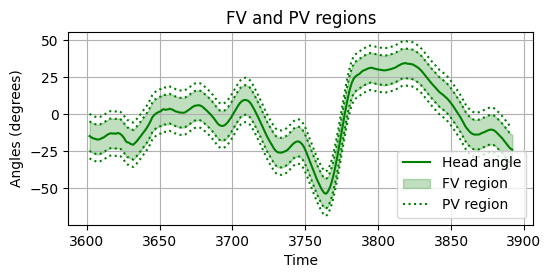}
  \caption{\small Driver's field of vision. The shaded area represents the FV region, while the area that extends up to the dotted line represents the driver's PV region.}
  \label{fig:margins}
\end{figure}

To mitigate the algorithm's limitations stemming from the absence of eye movement tracking, we defined two driver's gaze regions. Focus Vision (FV) ($10^{\circ}$ span) is the area where the driver focuses their vision to capture information from what they are observing. Peripheral Vision (PV) ($5^{\circ}$ span) is the region in which the driver can unconsciously perceive environmental changes. At a cognitive level, the information obtained in this region can alert the driver and change their focus of attention. These areas are depicted in Figure \ref{fig:margins}.
These values are taken from \cite{Ren2015}, where the authors explore the registration of drivers' eye movements and vehicle driving parameters while navigating left- and right-hand curves. The study findings indicated that drivers' gaze direction varies around the reference axis, with fixation points in the region surrounding the horizontal gaze. A driver must turn their head when the eye-yaw angle exceeds $15^{\circ}$ to perceive target size and position.

\subsection{Data Split}

Throughout the entire trajectory, our study specifically targeted a few locations for analyzing the attention of the safety driver. These locations were carefully selected to encompass scenarios where the AV needs to execute intricate maneuvers.

\begin{figure}[h]
  \centering
  \includegraphics[width=0.95\columnwidth]{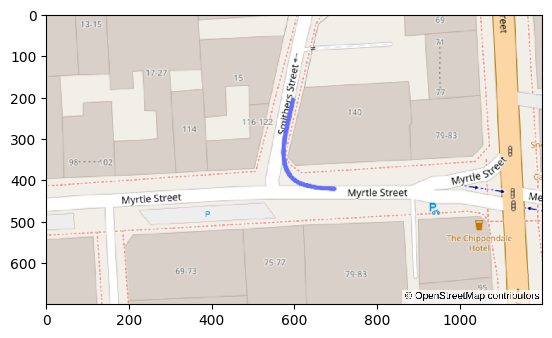}
  \caption{\small Intersection with a sample vehicle path in blue.}
  \label{fig:intersection}
\end{figure}

The main location evaluated corresponds to the intersection between Smithers St and Myrtle St, a T-junction, as depicted in Fig \ref{fig:intersection}.
This intersection was of special interest because the vehicle drives through a street with some traffic, and makes a turn to a perpendicular street with high traffic of both vehicles and pedestrians. In this way, the driver is forced to observe pedestrians and vehicles on both sides.
By examining the challenging locations, we aimed to gain valuable insights into the safety driver's attention during turning maneuvers.

\subsection{Attention Classification}

The objective of this process is to evaluate the level of attention exhibited by the safety driver towards the surrounding objects. To achieve this, we compute the intersection between the head orientation of the driver and the objects detected by the vehicle. Specifically, we use the driver's head yaw angle in conjunction with the calculated angles of the objects based on their distances. By intersecting these angles, we can determine whether the driver directed their attention towards the location of each object. This resulting intersection is subsequently compared to the regions of the FV and PV, enabling the derivation of a metric for classifying the driver's attention.

%This process aims to assess whether the safety driver is paying due attention to objects in their surroundings, which is achieved by computing the intersection between the angles of the objects and the safety driver's head. The resulting intersection is then compared to the regions of FV and PV to obtain a metric for classifying the driver's attention.

Objects within the FV region are considered the most relevant for analysis, while objects in the PV region are also taken into account with a 50\% weight compared to the former. This weight was determined empirically. This is demonstrated in \cite{Wolfe2017}, where authors assess visual attention within a limited portion of the visual field, specifically examining drivers' ability to attend to objects in the centre while being aware of those in the periphery. They suggest that, with the rise of automated vehicles, understanding the capabilities of the peripheral visual field becomes crucial for maintaining situational awareness. Thus, peripheral information may play a significant role in a driver's overall awareness and should be considered in the design of interfaces aiming to enhance this awareness.

\begin{figure*}[t]
\vspace{2mm}
  \centering
  \includegraphics[width=\textwidth]{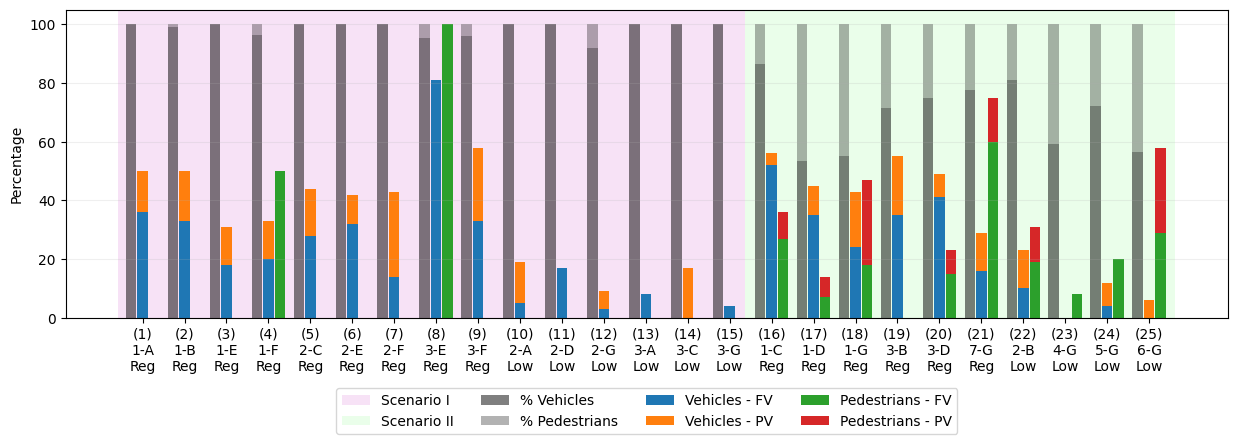}
  \caption{\small A breakdown of 25 cases in the analysis. Grey bars illustrate the ratio between the number of detected vehicles (depicted in dark grey) and pedestrians (in light grey) present in the scene. The blue-orange bars display the percentages of vehicles within the Focus Vision (FV) and Peripheral Vision (PV) regions, respectively. Likewise, the green-red bars represent the percentages of pedestrians situated in the FV and PV regions. The purple background is corresponding to the Scenario I cases and the yellow background corresponds to Scenario II. Cases have been ordered by scenarios and attention level for a better understanding. The bottom axis shows the case reference number (between parenthesis), the second row shows the number of the lap (1-7) driven by each supervisor (A-G), and the third row is the attention level}
  \label{fig:attention-observed-objects-full}
\end{figure*}
The chosen driving area generally has a higher number of vehicles compared to pedestrians. However, since there is no vehicle mapping information available, it is difficult to determine the relevance of specific vehicles or pedestrians to the driver. Therefore, it is necessary to consider both stationary and moving vehicles, as well as pedestrians on sidewalks or crossing the street.
We using a percentage metric for evaluating observed objects at this intersection, as it might mitigate potential biases. By calculating the percentage based on the total number of detected objects, a more balanced assessment can be achieved.
In this way, we calculated which objects the driver observed and in which region of the driver's vision they were located. This information is used to classify the objects based on their relevance to the driver. In order to identify regular and low levels of attention, we use two K-Means classifiers to measure the driver's attention to both vehicles and pedestrians.

A third classifier evaluates the relationship between the observed vehicles and pedestrians in the scene, enabling the classification of situations into two groups: those where vehicles predominate, and those where the relationship between both is similar. We call them Scenario I and Scenario II, respectively. Cases within each scenario are shown in \ref{fig:attention-observed-objects-full}. The purple background is Scenario I and the green background is Scenario II. A detailed description of both scenarios is given in Section \ref{sec:Results}.
This facilitates the identification of potential areas where attention may be lacking.
Once the K-Means classifiers have evaluated the collected data, the driver's attention level is divided into regular or low. To accomplish this, the outputs from the two first classifiers mentioned are inputs into an additional K-Means classifier that combines the information to determine the driver's level of attention to their surroundings. This approach enables a comprehensive assessment of the driver's attention and more accurately determines their overall attention level.

\section{Results} \label{sec:Results}

\subsection{Quantitative Analysis}
A total of 25 cases were evaluated, with 10 falling into the low-attention category and 15 in the regular-attention category. Fig. \ref{fig:attention-observed-objects-full} illustrates the distinctive characteristics of each case, featuring three informative bars.
For a better interpretation, cases have been ordered by scenarios and driver attention.
The first bar showcases the percentage of participation of the detected objects within the scene. The dark grey segment represents the percentage of vehicles, while the light grey segment represents pedestrians. Together, these two segments account for 100\% of the detected objects.

The information collected allows for the identification of two distinct driving scenarios. Scenario I corresponds to situations where vehicles represent between 95\% and 100\% of the detected objects, with the remaining percentage, if any, attributed to pedestrians. As a result, a low pedestrian observation rate is expected in this scenario. Out of the cases evaluated, 15 fall into this category.
Scenario II, on the other hand, is characterized by a higher proportion of pedestrians, ranging from 5\% to 45\% of the observed objects. This scenario is represented by the remaining 10 cases.

The second bar provides insights into the percentage of vehicles observed by the driver. The blue segment indicates the proportion of vehicles observed within the FV region, while the orange segment represents the percentage observed within the PV region.
The third bar displays the percentage of observed pedestrians. The green segment corresponds to pedestrians observed within the FV region, while the red segment represents those within the PV region.

This information served as input for the initial set of classifiers. The outputs generated were subsequently utilized as inputs for the subsequent classifier, creating a cascaded approach. This sequential process allowed for the final classification of each case to be successfully conducted.

The cases categorized as low attention are marked by a low percentage of observed vehicles and pedestrians overall.
In the first scenario, the percentage of observed vehicles does not exceed 20\% in any of the cases, while the observation rate of pedestrians is nearly non-existent, as expected.
In the second scenario, low attention is characterized by a higher observation rate of pedestrians compared to vehicles. In the best-case scenario (column 25), the observed pedestrians represent less than 25\% of the total observations, with the relationship between detected vehicles and pedestrians being close to 50\%. Out of this percentage, half were observed in the FV region, while the other half were observed in the PV region.

Regular-attention cases are marked by an observation rate of vehicles ranging from 30\% to 60\% of those detected and pedestrian observation rates between 0\% and 25\%.
In Scenario I, the scarcity of pedestrians means that observations are typically very low or non-existent.
However, there are two cases in this scenario that deviate from this pattern, namely columns 4 and 8, where pedestrians were observed at high percentages of 50\% and 100\%, respectively.
Upon closer examination, it was found that in the column-8 case, the vehicle made only one detection corresponding to a pedestrian, which suggests a potential failure in the detection of the object. Typically, multiple detections of the same object are observed over time when the object is successfully detected, as shown in Figure \ref{fig:position}. In this case, the absence of such multiple detections means that the pedestrian's actual presence cannot be guaranteed. If the pedestrian were truly present, there would have been several nearby detections corresponding to the same object.
Similarly, in the column-4 case, only one of the two pedestrian detections in the scene accounted for 50\% of the driver's observations. These detections were separated by a time of 71.56 s, making it unlikely for them to be the same pedestrian, since the vehicle was not stationary during that time.
Despite these potential failures in detection, these cases were still correctly classified.

\begin{figure}[h]
\vspace{2mm}
  \centering
  \includegraphics[width=0.95\columnwidth]{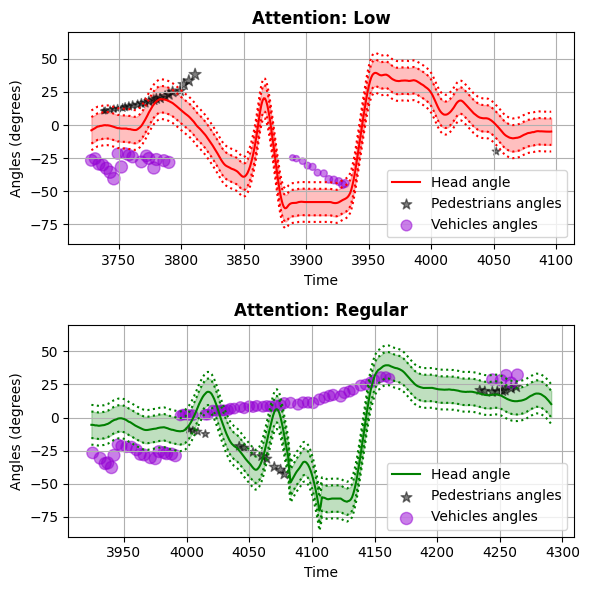}
  \caption{\small Samples for each attention classification. In the low-attention case, there is limited awareness of the objects present in the environment, which stands in stark contrast to the heightened observation associated with the high-attention case.}
  \label{fig:attention-examples}
\end{figure}

\begin{table} [h]
  \centering
  \begin{tabular}{ | c | c | c | c | c | } 
  \hline
  Attention & Veh. in FV & Veh. in PV & Ped. in FV & Ped. in PV \\
  \hline
  Low       & 0\%   & 6\%   & 29\%  & 29\% \\
  Regular   & 16\%  & 13\%  & 60\%  & 15\% \\  
  \hline
  \end{tabular}
  \caption{\small Veh. in FV: Vehicles observed by the driver in the FV region; Veh. in PV: Vehicles observed in the PV region; PDPed. in FV: Pedestrians observed by the driver in the FV region; Ped. in PV: Pedestrians observed in the PV region.}
  \label{tab:attention-examples}
\end{table}

Some samples of Scenario II and the corresponding attention classification are shown in Fig. \ref{fig:attention-examples}. For a better understanding of the examples, Table \ref{tab:attention-examples} shows the metrics for each case.

For the regular-attention sample, the safety driver observed 16\% of the vehicle detections within the FV region and 13\% of the vehicle detections within the PV region. This accounts for a total of 29\% of the vehicle detections present in the scene. Furthermore, the driver observed 60\% and 15\% of the pedestrian detections within the FV and PV regions, respectively, corresponding to 75\% of the total observed pedestrian detections.
Regarding the total number of objects detected, the percentage of vehicles observed is only 22.22\% and that of pedestrians represents 16.67\%.

For the low-attention, the driver exhibited a lower observation percentage for vehicle detections, with 0\% and 6\% observed within the FV and PV regions, respectively. They has observed 29\% of the pedestrians within FV region and 29\% within the PV region.
These percentages with respect to the total detected objects result in 3.64\% for vehicles and 25.45\% for pedestrians.

A significant finding arises from the analysis of drivers' attention during each lap regardless of which scenario it occurred in. As is shown in Table \ref{tab:drivers}, initially all drivers were classified as having regular attention in the first lap. However, as the subsequent laps unfolded, the patterns became more diverse. Out of the seven drivers, two (29\%) consistently maintained regular attention throughout all laps. Meanwhile, four drivers (57\%) experienced fluctuations between regular and low attention or low and regular attention. In one particular case (14\%), attention decreased and remained consistently low throughout the rest of the laps. This trend can be attributed to drivers gaining confidence in their vehicles, which may contribute to an overall decrease in attention levels.

\begin{table} [h]
  \centering
  \begin{tabular}{ | c | c | c | c | c | } 
  \hline
  Driver    & Lap 1     & Lap 2     & Lap 3     & Lap 4-7       \\
  \hline
  A         & Regular   & Low       & Low       & -             \\
  B         & Regular   & Low       & Regular   & -             \\
  C         & Regular   & Regular   & Low       & -             \\
  D         & Regular   & Low       & Regular   & -             \\
  E         & Regular   & Regular   & Regular   & -             \\
  F         & Regular   & Regular   & Regular   & -             \\
  G         & Regular   & Low       & Low       &  Low/Regular  \\
  \hline
  \end{tabular}
  \caption{\small Attention levels given by drivers on each lap driven.}
  \label{tab:drivers}
\end{table}

\begin{figure*}[t]
\centering
\vspace{2mm}
\begin{subfigure}[b]{0.485\textwidth}
   \includegraphics[trim={0 2cm 0 2cm},clip, width=1\linewidth]{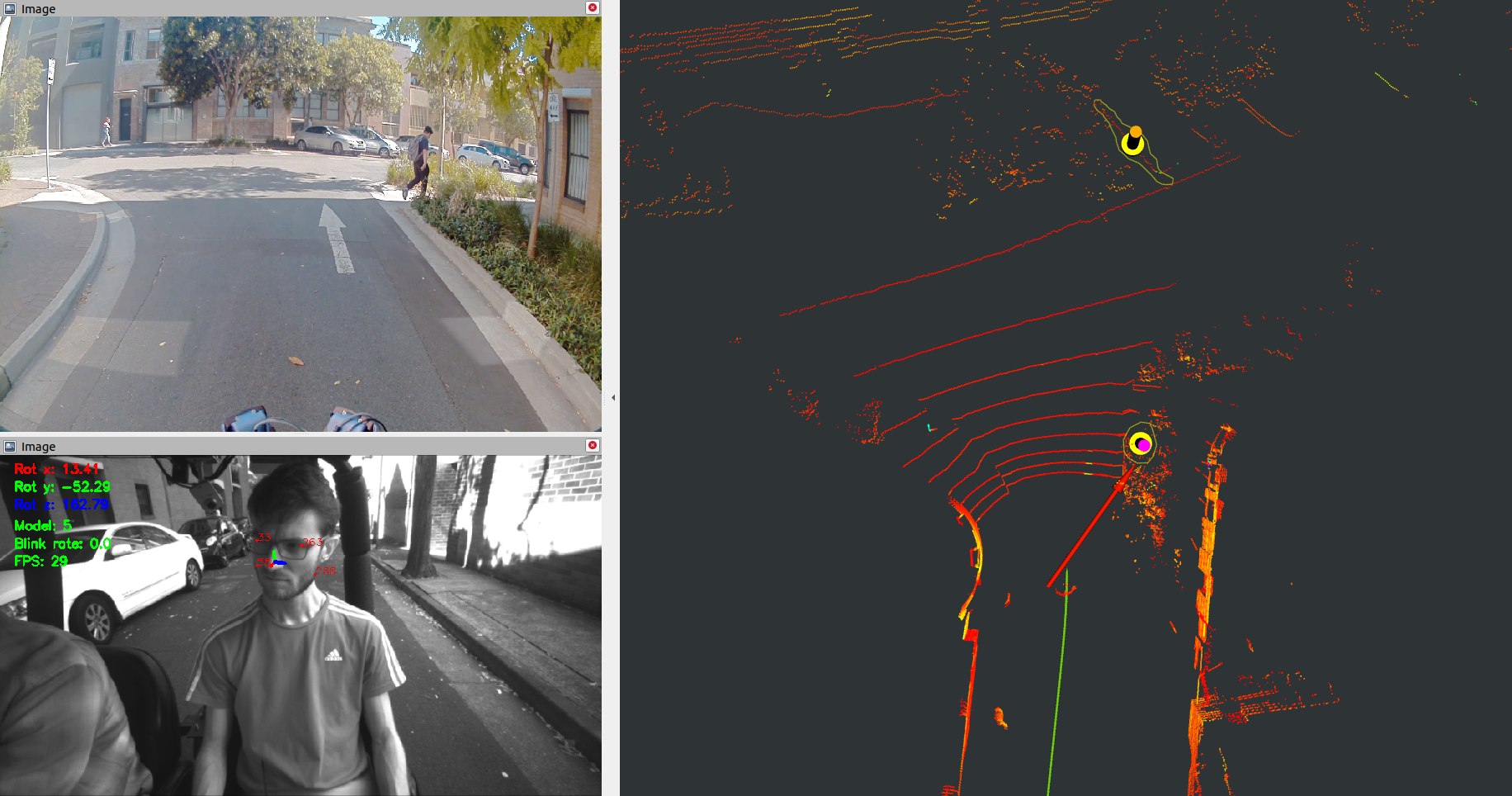}
   \caption{}
   \label{fig:sample18-a}
\end{subfigure}
\begin{subfigure}[b]{0.4\textwidth}
   \includegraphics[trim={0 0 0 0.7cm},clip,width=1\linewidth]{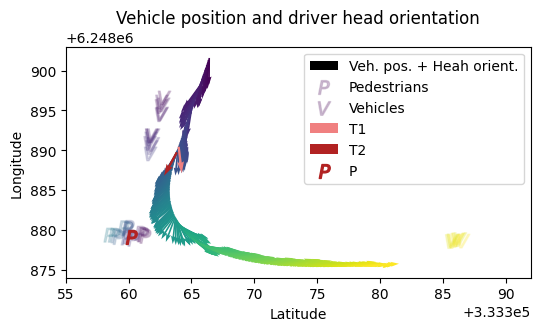}
   \caption{Vehicle position and driver head orientation}
   \label{fig:sample18-c} 
\end{subfigure}

%\enspace
\begin{subfigure}[b]{0.485\textwidth}
   \includegraphics[trim={0 2cm 0 2cm},clip,width=1\linewidth]{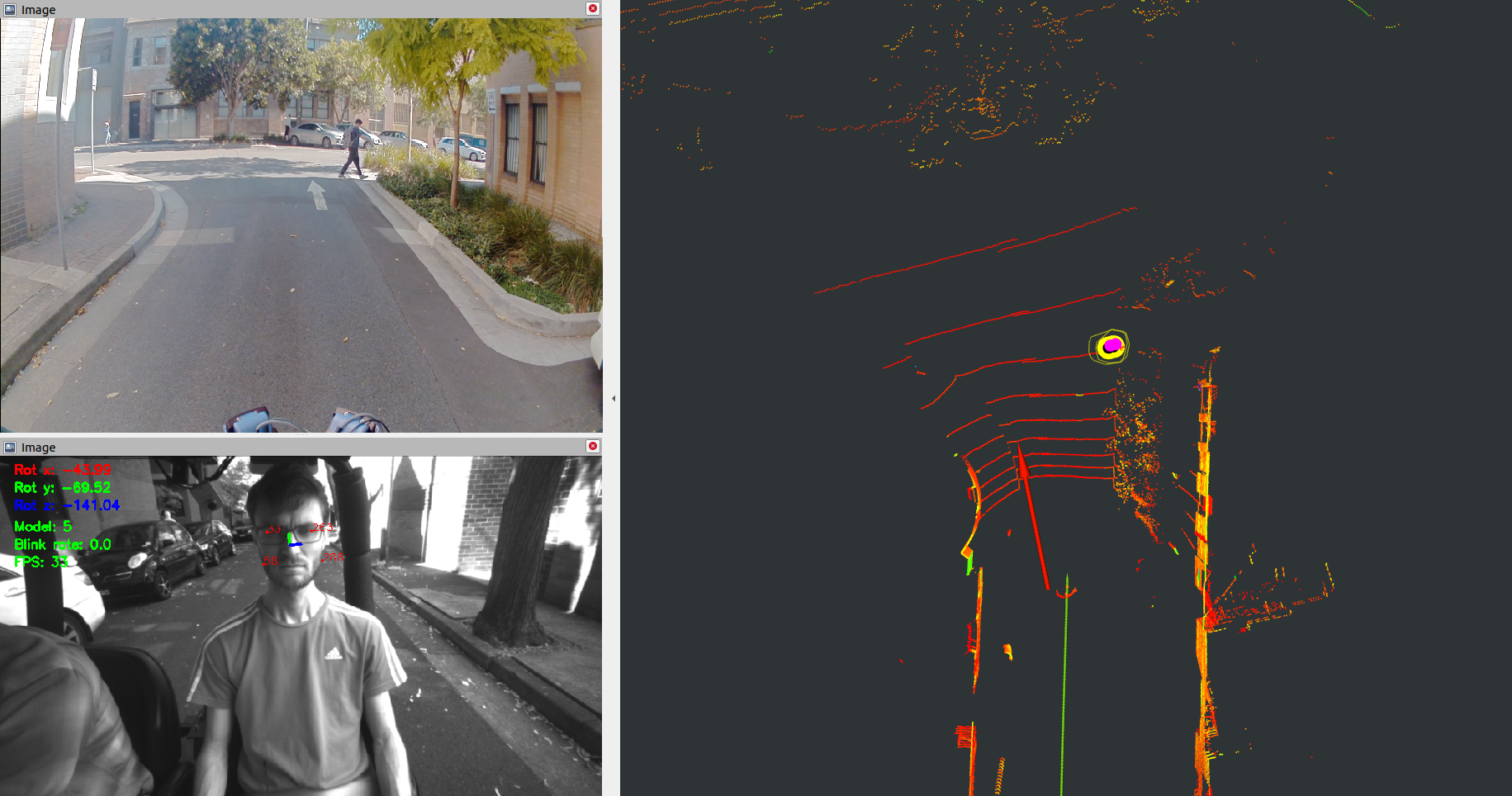}
   \caption{}
   \label{fig:sample18-b}
\end{subfigure}
%\enspace
\begin{subfigure}[b]{0.4\textwidth}
\centering
   \includegraphics[trim={0 0 0 0.7cm},clip,width=1\linewidth]{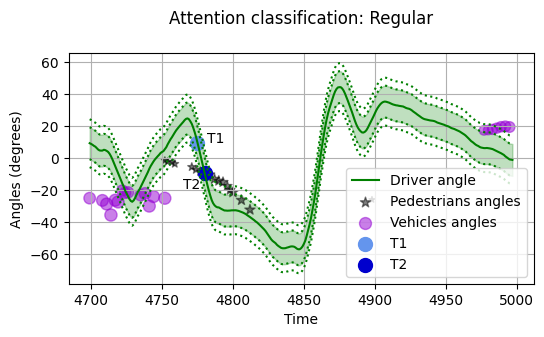}
   \caption{Attention classification: Regular}
   \label{fig:sample18-d} 
\end{subfigure}
\caption[]{\small Analysis of two moments of a sequence, T1 and T2. (a) and (c) show the pictures of the vehicle's frontal camera (left-top) showing the road, and the internal IR camera (left-down) showing the driver. The right image is the data captured by the vehicle's systems. (b) represents both times by a couple of red arrows. Also, the pedestrian detection corresponding to T2 is coloured red as well. (d) shows T1 and T2 as blue dots in the angle's plot}
\label{fig:sample18}
\end{figure*}

This behaviour can be observed in Fig. \ref{fig:attention-examples} which showcases the drivers and their respective laps displayed at the bottom of the bars. The top row represents the case reference number, followed by the corresponding lap numbers (1-7) of each driver and the driver identifiers A-G in the second row. The attention level for each case is indicated in the bottom row.

\subsection{Case Study}

The following case demonstrates how information from the environment works with the orientation of the driver's head, and how the angles of surrounding objects are used to intersect each other and understand where the driver was looking.

This regular-attention case is shown in Fig. \ref{fig:sample18}, which illustrates two moments from the sequence. At moment T1 (Fig. \ref{fig:sample18-a}), the vehicle detects a pedestrian that the driver is not observing. However, 1.67 seconds later (T2), the driver turns their head and looks at the pedestrian (Fig. \ref{fig:sample18-c}).

Figs \ref{fig:sample18-a} and \ref{fig:sample18-c} exhibit information captured by the vehicle at both times. The top-left image shows the vehicle's frontal camera view, enabling vehicle and pedestrian recognition.
This is one of the three front cameras that the vehicle has and that were used to detect objects.
The bottom-left picture displays the IR camera view of the safety driver, featuring landmarks mentioned in Section \ref{sec:HeadOrientation} (red points), and orientation angles provided by three coloured lines close to the driver's nose. Additionally, some information the algorithm provides is shown on the top-left side.
The right image shows lidar information, with a red arrow indicating the driver's head orientation, a green path showing the vehicle's path, and a yellow-purple marker indicating the detected pedestrian.
In this can also be observed the detection of a vehicle in figure \ref{fig:sample18-a} that later stopped being detected in figure \ref{fig:sample18-b}.

This can be observed in the plot presented in Fig. \ref{fig:sample18-b}. This shows the overlap between the vehicle's position and the driver's head orientation.
The ``V'' and ``P'' markers in Fig. \ref{fig:sample18-b} represent the vehicles and pedestrians detected at the scene, respectively. Each group of vehicle detections corresponds to a different vehicle, while the group of pedestrian detections corresponds to the pedestrian observed in the camera.
The vehicle that was detected in Figure \ref{fig:sample18-a} and not in Figure \ref{fig:sample18-c} is not shown in this figure as it has been removed as part of the filtering mentioned in Section \ref{sec:filtering}.
The head orientation at both times is indicated by additional red arrows, corresponding to the blue dots in Fig. \ref{fig:sample18-d}. The detected pedestrian from the T2 observation is coloured in red.

In Fig. \ref{fig:sample18-d} the intersection between the head orientation and the pedestrian's angles with respect to the vehicle are shown. Although both situations presented correspond to the times in the blue points, in this figure is easy to see the driver has observed the parked vehicles before and even the pedestrian. In the general description of the situation, the driver paid attention to most of the detected objects in the scene and also followed the pedestrian for some time after T1. Mainly they were observed within the FV, meaning a better attention focus. It is why this case has been classified as a regular attention case.

\section{Conclusions and Future Work}

This research paper presents the findings of a classification study conducted on various real-life driving scenarios, focusing on distinguishing between regular and low driver attention. The attention metric we developed was validated using a dataset gathered from individuals acting as safety drivers in AV. Through this validation process, we observed a correlation between our proposed metric and the phenomenon of ``automation complacency''. Specifically, we found that individuals exhibited higher levels of attention during the first lap than in subsequent laps.

%It is important to highlight that due to the experimentation taking place in genuine traffic conditions, each scene was unique and non-repeatable. Moreover, the driving environment influenced the driver's monitoring approach, preventing the establishment of a consistent pattern of head movements as a reliable indicator of attention. Additionally, multiple drivers were evaluated, making it impossible to compare the results against a definitive benchmark.

%Given these constraints, 
The methodology employed in this study involved leveraging information from the surrounding environment of the vehicle in conjunction with the driver's head movements to identify observed objects within the environment. This allowed for inferring the driver's attention based on whether specific objects were observed during driving.
By considering vehicle and environmental data alongside driver information, the proposed system is well-positioned to offer comprehensive metrics. These metrics play a crucial role in facilitating appropriate warnings and promoting safe driving behaviors. 
%To accomplish this, a set of classifiers was employed, which received input regarding the detected vehicles and pedestrians by the ego-vehicle, as well as those observed by the driver.

%By utilizing this approach, two distinct scenarios were identified based on the relationship between the detected vehicles and pedestrians, with the information solely originating from the sensors of the ego-vehicle.
%The levels of driver attention were determined by combining the information from the vehicle with the driver's head orientation. The head orientation played a crucial role in recognizing the objects observed during driving.
%Of the two attention levels determined, the regular attention level is characterized by observing an average of up to 50\% of the vehicles and 20\% of the pedestrians in the scene. Conversely, the low attention level occurs when the driver's observations decrease to an average of 10\% for both vehicles and pedestrians.

%Considering vehicle and environmental data in conjunction with driver information, the proposed system is better equipped to provide metrics that serve to enable appropriate warnings to encourage safe driving behaviours. This dual-source approach offers a comprehensive solution to monitoring driver attentiveness in dynamic driving environments.

Our future plans involve expanding our dataset and refining the processing techniques to obtain more pertinent data from the vehicle's surroundings. To achieve this goal, we are working towards incorporating map information into our system. This additional data will provide us with a more comprehensive understanding of the environment. With the inclusion of map information, we will be able to differentiate between pedestrians on the sidewalk and those crossing the street or about to do so. Additionally, we can identify right-of-ways, determine the direction of vehicle travel, and consider other relevant factors. 
By implementing this system, we have the potential to significantly improve overall driving safety by providing metrics that generate alerts to drivers when it is determined that their attention to their surroundings is inadequate. %A low-level warning will be issued when the driver performs actions that may indicate potential distraction. If the low-level warning goes unaddressed or if the driver engages in risky behaviors, a high-level warning will be triggered. These warnings will only be reset when the driver adopts and consistently maintains a safe condition for a specified period of time. This safe condition is typically characterized by an increased level of attention to environmental obstacles.}

%\IEEEtriggeratref{7}

\bibliographystyle{IEEEtran}
\bibliography{bibliography}

\begin{thebibliography}{10}
\providecommand{\url}[1]{#1}
\csname url@samestyle\endcsname
\providecommand{\newblock}{\relax}
\providecommand{\bibinfo}[2]{#2}
\providecommand{\BIBentrySTDinterwordspacing}{\spaceskip=0pt\relax}
\providecommand{\BIBentryALTinterwordstretchfactor}{4}
\providecommand{\BIBentryALTinterwordspacing}{\spaceskip=\fontdimen2\font plus
\BIBentryALTinterwordstretchfactor\fontdimen3\font minus
  \fontdimen4\font\relax}
\providecommand{\BIBforeignlanguage}[2]{{%
\expandafter\ifx\csname l@#1\endcsname\relax
\typeout{** WARNING: IEEEtran.bst: No hyphenation pattern has been}%
\typeout{** loaded for the language `#1'. Using the pattern for}%
\typeout{** the default language instead.}%
\else
\language=\csname l@#1\endcsname
\fi
#2}}
\providecommand{\BIBdecl}{\relax}
\BIBdecl

\bibitem{Eriksson2017-1}
\BIBentryALTinterwordspacing
A.~Eriksson and N.~A. Stanton, ``Takeover time in highly automated vehicles:
  Noncritical transitions to and from manual control,'' \emph{Human Factors},
  vol.~59, no.~4, pp. 689--705, 2017, pMID: 28124573. [Online]. Available:
  \url{https://doi.org/10.1177/0018720816685832}
\BIBentrySTDinterwordspacing

\bibitem{Eriksson2017-2}
\BIBentryALTinterwordspacing
A.~Eriksson and N.~Stanton, ``Driving performance after self-regulated control
  transitions in highly automated vehicles,'' \emph{Human Factors}, vol.~59,
  no.~8, pp. 1233--1248, 2017, pMID: 28902526. [Online]. Available:
  \url{https://doi.org/10.1177/0018720817728774}
\BIBentrySTDinterwordspacing

\bibitem{FlynnEvans2021}
E.~Flynn-Evans, L.~Wong, Y.~Kuriyagawa, N.~Gowda, P.~Cravalho, S.~Pradhan,
  N.~Feick, N.~Bathurst, Z.~Glaros, T.~Wilaiprasitporn, K.~Bansal, J.~Garcia,
  and C.~Hilditch, ``Supervision of a self-driving vehicle unmasks latent
  sleepiness relative to manually controlled driving,'' \emph{Scientific
  Reports}, vol.~11, 09 2021.

\bibitem{Pipkorn2022}
\BIBentryALTinterwordspacing
L.~Pipkorn, M.~Dozza, and E.~Tivesten, ``Driver visual attention before and
  after take-over requests during automated driving on public roads,''
  \emph{Human Factors}, 2022, pMID: 35708240. [Online]. Available:
  \url{https://doi.org/10.1177/00187208221093863}
\BIBentrySTDinterwordspacing

\bibitem{Koopman2019}
P.~Koopman and B.~Osyk, ``Safety argument considerations for public road
  testing of autonomous vehicles,'' 04 2019.

\bibitem{Tran2021}
\BIBentryALTinterwordspacing
N.~H. Tran and A.~C. Nayak, ``Self-driving cars and driver alertness,'' 2021.
  [Online]. Available: \url{https://arxiv.org/abs/2107.14036}
\BIBentrySTDinterwordspacing

\bibitem{Cunningham2018}
\BIBentryALTinterwordspacing
M.~L. Cunningham and M.~A. Regan, ``Driver distraction and inattention in the
  realm of automated driving,'' \emph{IET Intelligent Transport Systems},
  vol.~12, no.~6, pp. 407--413, 2018. [Online]. Available:
  \url{https://ietresearch.onlinelibrary.wiley.com/doi/abs/10.1049/iet-its.2017.0232}
\BIBentrySTDinterwordspacing

\bibitem{Amina2020}
\BIBentryALTinterwordspacing
A.~Guettas, S.~Ayad, and O.~Kazar, ``Driver state monitoring system: A
  review,'' in \emph{Proceedings of the 4th International Conference on Big
  Data and Internet of Things}, ser. BDIoT'19.\hskip 1em plus 0.5em minus
  0.4em\relax New York, NY, USA: Association for Computing Machinery, 2020.
  [Online]. Available: \url{https://doi.org/10.1145/3372938.3372966}
\BIBentrySTDinterwordspacing

\bibitem{Kumar2018}
S.~Kumar, A.~Kalia, and A.~Sharma, ``Predictive analysis of alertness related
  features for driver drowsiness detection,'' in \emph{Intelligent Systems
  Design and Applications}, A.~Abraham, P.~K. Muhuri, A.~K. Muda, and
  N.~Gandhi, Eds.\hskip 1em plus 0.5em minus 0.4em\relax Cham: Springer
  International Publishing, 2018, pp. 368--377.

\bibitem{Friedrichs2020}
F.~Friedrichs, ``Driver alertness monitoring using steering, lane keeping and
  eye tracking data under real driving conditions,'' 2020.

\bibitem{JAMSON2013116}
\BIBentryALTinterwordspacing
A.~H. Jamson, N.~Merat, O.~M. Carsten, and F.~C. Lai, ``Behavioural changes in
  drivers experiencing highly-automated vehicle control in varying traffic
  conditions,'' \emph{Transportation Research Part C: Emerging Technologies},
  vol.~30, pp. 116--125, 2013. [Online]. Available:
  \url{https://www.sciencedirect.com/science/article/pii/S0968090X13000387}
\BIBentrySTDinterwordspacing

\bibitem{mental}
M.~Recarte and L.~Nunes, ``Mental workload while driving: Effects on visual
  search, discrimination, and decision making,'' \emph{Journal of experimental
  psychology. Applied}, vol.~9, pp. 119--37, 07 2003.

\bibitem{usyd_dataset}
W.~Zhou, J.~S. Berrio, C.~De~Alvis, M.~Shan, S.~Worrall, J.~Ward, and E.~Nebot,
  ``Developing and testing robust autonomy: The university of sydney campus
  data set,'' \emph{IEEE Intelligent Transportation Systems Magazine}, vol.~12,
  no.~4, pp. 23--40, 2020.

\bibitem{wang2022}
C.-Y. Wang, A.~Bochkovskiy, and H.-Y.~M. Liao, ``{YOLOv7}: Trainable
  bag-of-freebies sets new state-of-the-art for real-time object detectors,''
  \emph{arXiv preprint arXiv:2207.02696}, 2022.

\bibitem{Mediapipe2019-1}
\BIBentryALTinterwordspacing
C.~Lugaresi, J.~Tang, H.~Nash, C.~McClanahan, E.~Uboweja, M.~Hays, F.~Zhang,
  C.-L. Chang, M.~G. Yong, J.~Lee, W.-T. Chang, W.~Hua, M.~Georg, and
  M.~Grundmann, ``Mediapipe: A framework for building perception pipelines,''
  2019. [Online]. Available: \url{https://arxiv.org/abs/1906.08172}
\BIBentrySTDinterwordspacing

\bibitem{Mediapipe2019-2}
\BIBentryALTinterwordspacing
Y.~Kartynnik, A.~Ablavatski, I.~Grishchenko, and M.~Grundmann, ``Real-time
  facial surface geometry from monocular video on mobile gpus,'' 2019.
  [Online]. Available: \url{https://arxiv.org/abs/1907.06724}
\BIBentrySTDinterwordspacing

\bibitem{Collins2014}
T.~Collins and A.~Bartoli, ``Infinitesimal plane-based pose estimation,''
  \emph{International Journal of Computer Vision}, vol. 109, pp. 252--286,
  2014.

\bibitem{article_yolo}
C.~Guo, X.-l. Lv, Y.~Zhang, and M.-l. Zhang, ``Improved yolov4-tiny network for
  real-time electronic component detection,'' \emph{Scientific Reports},
  vol.~11, 11 2021.

\bibitem{paper:VoMa2006}
B.-T. Vo and W.-K. Ma, ``The {G}aussian mixture probability hypothesis density
  filter,'' \emph{IEEE Transactions on Signal Processing}, vol.~54, no.~11, p.
  4091–4104, Nov. 2006.

\bibitem{Ren2015}
Y.-y. Ren, X.-s. Li, X.-l. Zheng, Z.~Li, and Q.-c. Zhao, ``Analysis of
  drivers’ eye-movement characteristics when driving around curves,''
  \emph{Discrete Dynamics in Nature and Society}, vol. 2015, pp. 1--10, 03
  2015.

\bibitem{Wolfe2017}
\BIBentryALTinterwordspacing
B.~Wolfe, J.~Dobres, R.~Rosenholtz, and B.~Reimer, ``More than the useful
  field: Considering peripheral vision in driving,'' \emph{Applied Ergonomics},
  vol.~65, pp. 316--325, 2017. [Online]. Available:
  \url{https://www.sciencedirect.com/science/article/pii/S0003687017301631}
\BIBentrySTDinterwordspacing

\end{thebibliography}

\end{document}